\documentclass[10pt,twocolumn,letterpaper]{article}

\usepackage{times}
\usepackage{epsfig}
\usepackage{graphicx}
\usepackage{amsmath}
\usepackage{amssymb}
\usepackage{algorithmic}
\usepackage[ruled]{algorithm2e}
\usepackage{makecell}
\usepackage{subcaption}

\usepackage[pagebackref=true,breaklinks=true,letterpaper=true,colorlinks,bookmarks=false]{hyperref}

\begin{document}

\title{Deep Joint Face Hallucination and Recognition}

\author{Junyu Wu\\
	Sun Yat-sen University\\
	{\tt\small wujunyu2@mail2.sysu.edu.cn}
	\and
	Shengyong Ding\\
	Sun Yat-sen University\\
	{\tt\small 1633615231@qq.com}
	\and
	Wei Xu\\
	Sun Yat-sen University\\
	{\tt\small xuwei1993@qq.com}
	\and
	Hongyang Chao\\
	Sun Yat-sen University\\
	{\tt\small isschhy@mail.sysu.edu.cn}
}

\maketitle

\begin{abstract}

Deep models have achieved impressive performance for face hallucination tasks.
However, we observe that directly feeding the hallucinated facial images into recognition models
can even degrade the recognition performance despite the much better visualization quality.
In this paper, we address this problem by jointly learning a deep model for two tasks, i.e. face
hallucination and recognition. In particular, we design an end-to-end deep convolution network with
hallucination sub-network cascaded by recognition sub-network. The recognition sub-network are responsible for
producing discriminative feature representations using the hallucinated images as inputs
generated by hallucination sub-network. During training, we feed LR facial images into the
network and optimize the parameters by minimizing two loss items, i.e. 1) face hallucination loss
measured by the pixel wise difference between the ground truth HR images and network-generated
images; and 2) verification loss which is measured by the classification error and intra-class distance.
We extensively evaluate our method on LFW and YTF datasets. The experimental results show that
our method can achieve recognition accuracy $97.95\%$ on 4x down-sampled LFW testing set, outperforming
the accuracy $96.35\%$ of conventional face recognition model.
And on the more challenging YTF dataset, we achieve recognition
accuracy $90.65\%$, a margin over the recognition accuracy $89.45\%$ obtained by conventional face recognition model
on the 4x down-sampled version.

\end{abstract}

\section{Introduction}

\begin{figure}[!htb]
	\begin{center}
		\includegraphics [width=0.4\textwidth]{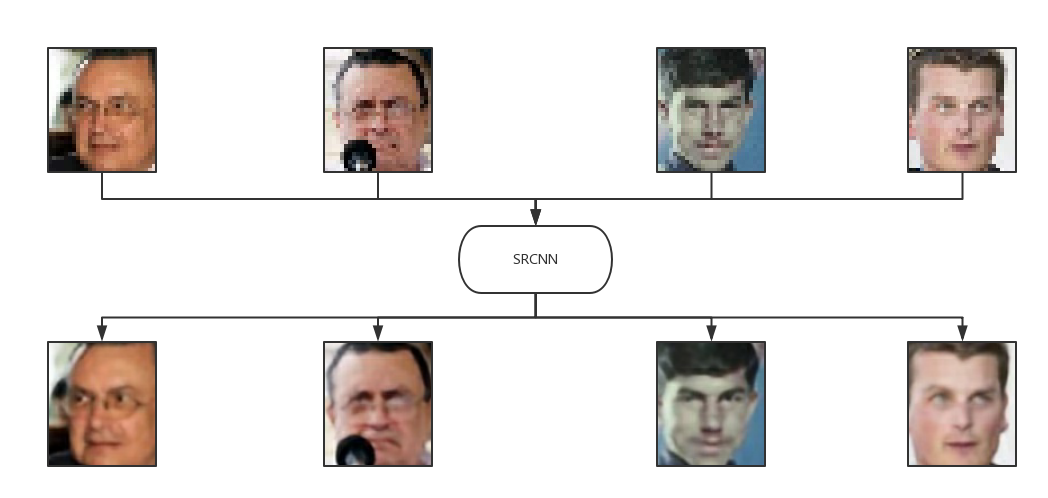}
		\caption{Images generated by SRCNN.}
		\label{fig:srexample}
	\end{center}
\end{figure}

Face hallucination and recognition are critical components for a lot of applications, e.g.
law enforcement and video surveillance. Face hallucination aims at producing HR (high-resolution) facial
images from LR (low-resolution) images \cite{DBLP:journals/ijcv/LiuSF07}. Face recognition targets at verifying whether two
facial images are from the same identity by designing discriminative features and similarities \cite{turk1991eigenfaces}.
Empirical studies \cite{Lui2009A} in face recognition proved that a minimum face resolution
between $32 \times 32$ and $64 \times 64$ is required for stand-alone recognition algorithms.
\cite{DBLP:journals/pami/TorralbaFF08} reported a significant performance drop when the image resolution is decreased
below $32 \times 32$ pixels.
It is natural to expect that hallucinated face images can improve the recognition performance
for LR facial images. Unfortunately, we find that this expectation does not hold in a
lot of cases. As an example, Figure \ref{fig:srexample} shows typical LR versions of LFW \cite{LFWTech} and
its hallucinated counterparts generated by SRCNN \cite{srcnn}. We can clearly see that hallucinated
versions have much better details and sharpness. However, feeding the hallucinated versions to a state-of-the-art
recognition model can even degrade the recognition performance compared with the LR versions
(from $96.35\%$ to $96.30\%$).

\cite{DBLP:conf/pricai/ZhangGLC16} reported similar conclusion: SR algorithms may perform poorly on recognition task since SR algorithms focus more on visual enhancement rather than classification accuracy.
Considering the SR model and recognition model are trained separately, this phenomenon is not hard to be
explained as each model has no signals or feedbacks from the other one during the training. Thus we propose
a novel method to jointly optimize these two models under a unified convolutional neural network.
Our Joint Model is based on an end-to-end CNN which can be seen as composed of two sub-networks, i.e. hallucination sub-network
followed by recognition sub-network. During testing, one LR image is fed to the end-to-end network so that the hallucination
sub-network produces a hallucinated facial image (as intermediate feature maps). Then this
hallucinated image is fed to the recognition sub-network to generate a representation vector for recognition.
In order to jointly solve these two tasks, LR face images are provided with its HR versions
as well as their identities in the training stage. With these enriched training samples,
we introduce two loss items to solve the parameters, i.e. hallucination loss and recognition loss.
The hallucination loss is defined as the squared difference between the generated image and ground truth HR.
The recognition loss follows the recently published literature \cite{centerloss} which is defined as the weighted
sum of classification error and intra-class distance (the distance between each sample and its center in
the feature space). Intuitively, classification error is to separate different classes as far as possible
while the intra-class distance is to shrink the samples of one class.

To the best of our knowledge, there are few works studying the joint learning of hallucination and recognition
for face images. The most similar work to ours is proposed by Z Wang et al \cite{studyvlr}. In this work, the authors first
train a SR network. Then two fully-connected layers are stacked on this pretrained SR network
to learn a classification model. During the learning of this classification model, the super
resolution loss is not applied anymore, i.e. SR module only acts as pretraining rather than joint supervision.
In contrast to this work, we focus on face domain and extensively study the joint effect of SR and
recognition using state-of-the-art network architectures to rigorously evaluate the improvements brought by the Joint Model.

We extensively evaluate our method on public dataset, i.e. LFW and the YTF. We obtain a set of models for
thorough comparison to demonstrate the effect of the Joint Model. Our experimental results show
that the result of Joint Model outperforms the independently trained models by a margin of $0.63\%$ on LFW.

In summary, our contributions are mainly two folded:
\begin{itemize}
	\item A joint end-to-end model which simultaneously solve hallucination task and recognition task.
	\item Extensive performance reports of hallucination and recognition performance on facial dataset.
\end{itemize}

\section{Related Work}
The related work to our method can be roughly divided into 3 groups as follows.

\subsection{Face Recognition}
The shallow models, e.g. Eigen face \cite{turk1991eigenfaces}, Fisher Face \cite{belhumeur1997eigenfaces},
and Gabor based LDA \cite{liu2002gabor}, and LBP based LDA \cite{li2007illumination} usually rely on
handcrafted features and are evaluated on early datasets in controlled environments.
Recently, a set of deep face models have been proposed and greatly advanced the progress
\cite{taigman2014deepface,sun2014deep,yi2014learning,DBLP:FaceNet}. DeepID
\cite{sun2014deep} uses a set of small networks with each network observing a patch of the face
region for recognition. FaceNet \cite{DBLP:FaceNet} is another deep face model proposed recently, which are
trained by relative distance constraints with one large network. Using a huge dataset, FaceNet achieves
99.6\% recognition rate on LFW. \cite{DBLP:conf/eccv/WenZL016} proposed a loss function (called center loss) to minimize
the intra-class distances of the deep features, and achieved 99.2\% recognition rate on LFW using web-collected
training data.

\subsection{Super Resolution and Face Hallucination}
A category of state-of-art SR approaches
\cite{DBLP:journals/ijcv/FreemanPC00,DBLP:conf/cvpr/ChangYX04,DBLP:conf/cvpr/YangWHM08}
learn a mapping between LR / HR patches.
There have been some studies of using deep learning techniques for SR \cite{srcnn} \cite{vdsr}.
SRCNN \cite{srcnn} is a representative state-of-art method for deep learning based SR approach, which directly
models HR images with 3 layers: patch extraction / representation, non-linear mapping,
and reconstruction. \cite{vdsr} proposed a Very Deep Super-Resolution convolutional network, modeling high frequency information with a 20 weighted $3 \times 3$ layers network.

Conventional hallucination methods \cite{DBLP:conf/fgr/BakerK00,DBLP:journals/tsmc/WangT05} are often designed for controlled settings and cannot handle varying conditions.
Deep models are also applied to face hallucination tasks \cite{fhwild,DBLP:conf/eccv/ZhuLLT16}.
\cite{fhwild} proposed a Bi-channel Convolutional Neural Network, which extracts robust face representations from raw input
by using deep convolutional network, then adaptively integrates 2 channels of information to predict the HR image.

\subsection{Low Resolution Face Recognition}
Low-resolution face recognition (LR FR) aims to recognize faces from small size or poor quality
images with varying pose, illumination, expression, etc.
\cite{verylrfrproblem} reported a degradation of the recognition performance when face regions
became smaller than $16 \times 16$. \cite{studyvlr} proposed a Partially
Coupled Super-Resolution Networks (PCSRN), as the pre-training part of recognition model.

\section{Joint Model}
\begin{figure*}[!htb]
	\begin{center}
		\includegraphics [width=0.8\textwidth]{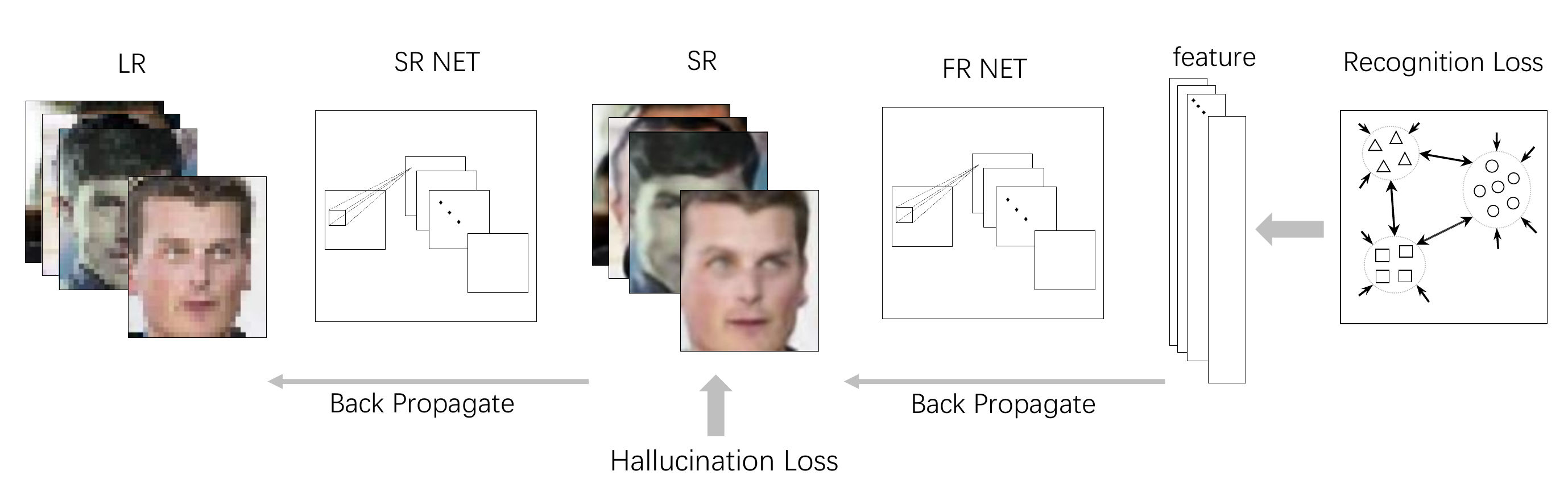}
		\caption{Illustration of the Joint Model.}
		\label{fig:DARI}
	\end{center}
\end{figure*}

We use one end-to-end network to jointly solve face hallucination and recognition.
Figure \ref{fig:DARI} illustrates the overall principle. This network consists of two parts, i.e. face
hallucination layers and recognition layers, which will be abbreviated as SRNET and FRNET respectively for convenience. In testing stage, the hallucination layers produce
a high resolution facial image ${I^{h}}$ for a low resolution facial image ${I^{l}}$.
The recognition layers then generates face representations ${x}$ using ${I^{h}}$ as input
which serves face recognition task. As these two parts are cascaded, these two steps will be
executed by one forward propagation, i.e. in end-to-end fashion.

An intuitive approach to implement this end-to-end network is cascade one well trained SRNET and one FRNET. However, as we aforementioned, such direct cascading will even degrade the overall recognition performance despite the output of the SRNET has better visualization and PSNR since the well trained FRNET has never seen samples generated by SRNET. 

In order to address this problem, we propose to jointly optimize these two networks so that each network can benefit from the other one. Figure \ref{fig:DARI} illustrates the overall principle. Given a set of low resolution facial images $I_i^l$ with their high resolution versions $\tilde I_i^h$ and the labels of identities $c_i$, the end-to-end model produces predicted high resolution facial images $I_i^h$ by SRNET and feature vectors $x_i$ by FRNET. This end-to-end network is jointly optimized so that $I_i^h$ are as close as possible to $\tilde I_i^h$ and $x_i$ should be able to separate different identities in the feature space. These two constraints can be further formulated as two loss items $L_h$ and $L_r$ in the overall objective function $L$ as follows where $W_h$ denotes the parameter set of SRNET and $W_r$ denotes the parameter set of FRNET with $\alpha$ and $\beta$ controls the weight of these two items:

\begin{equation}
L=\alpha L_h(I_i^l;W_h)+\beta L_r(I_i^l;W_h,W_r)
\end{equation}

Note $L_r$ depends on both parameter set $W_r$ and $W_h$ as FRNET uses the outputs of SRNET as inputs.
For the loss item $L_h$, we use the pixel wise difference between $I_i^h$ and $\tilde I_i^h$ as below: 

\begin{equation}
L_h=\Sigma_{i=1}^{m}||I_i^h-\tilde I_i^h||^2
\end{equation}

And for the recognition loss, we want to obtain representations that can discriminate different identities in the feature space under some similarity measure. We follow the recently published method, named center loss to model this constraint. In particular, this loss includes two items, i.e. classification error $L_c$ and the center loss $L_d$ which is defined as the mean intra-class distance between the samples and their centers.
We use $W_j$ to denote the $j$th column of the softmax weight matrix $W$  and $b_j$ for the bias terms, then $L_c$ can be defined as below where $n$ is the number of training samples: 
\begin{equation}
L_c=-\Sigma_{i=1}^{n} \log \frac{e^{W_{c_i} x_i+b_{c_i}}}{\Sigma_{j=1}^{n}e^{W_{j} x_{j}+b_{j}}}
\end{equation}

By using  $m_{c_i}$ to represent the center of class $c_i$ in the feature space, $L_d$ is then defined as:
\begin{equation}
L_d=\Sigma_{i=1}^{n}||x_i-m_{c_i}||^2
\end{equation}

In order to balance the softmax loss $L_c$ and center loss $L_d$, we can introduce weight parameters $\alpha$, $\beta_1$ and $\beta_2$ and define the overall loss function as:
\begin{equation}
L= \alpha L_h+\beta_1 L_c+\beta_2 L_d
\end{equation}

In the next section, we will give a method to solve this model in the end-to-end fashion.

\subsection{Optimization}
In this section, we show how to jointly solve our end-to-end model. As the softmax and center-loss are introduced, we use $W_s$ and $M=[m_1, m_2,..., m_k]$ to denote the softmax parameter set and center vectors of $k$ classes and give a parameterized version of the loss function to show the dependency of different items on the parameter set. 
\begin{equation}
\begin{aligned}
L(I_i^l;W_h,W_r,W_s,M) = {} & \alpha L_h(I_i^l;W_h) \\
& + \beta_1 L_c(I_i^l;W_h,W_r,W_s) \\
& +\beta_2 L_d(I_i^l;W_h,W_r,M)
\end{aligned}
\end{equation}

Due to the non-convexity of the loss function, we apply gradient descent algorithm to find the local minimum, i.e. calculating the gradient $\nabla W=[\nabla W_h \nabla W_r \nabla W_s]$ and update $W$ by this gradient with a learning rate iteratively. Note the update of $M$ is replaced by an approximate mechanism as adopted in literature \cite{centerloss} rather than the gradient method.

\textbf{Graident with respect to $W_r$ ($\nabla W_r$): } This gradient is relatively simple and can be obtained by running the standard back propagation algorithm after we calculate $\frac{\partial L}{\partial x_i}$ as the following chain rule holds:
\begin{equation}
\frac{\partial L}{\partial W_r}=\Sigma_{i=1}^{m}\frac{\partial L}{\partial x_i}\frac{\partial x_i}{\partial W_r}
\end{equation}

Actually, $\frac{\partial L}{\partial x_i}$ involves two terms according to the definition as below:

\begin{equation}
\frac{\partial L}{\partial x_i}=\beta_1 \frac{\partial L_c}{\partial x_i} + \beta_2 \frac{\partial L_d}{\partial x_i}
\end{equation}

The first term is rather simple according to the definition of $L_c$. However, the second term is a little bit complicated as $L_d$ depends on class center $m_j$ which further depends on $x_i$. In order to simplify the optimization algorithm, we follow the approach in literature \cite{centerloss}, i.e. fixing the center $m_j$ during calculation of $\frac{\partial L_d}{\partial x_i}$. This simplification gives us:
\begin{equation}
\frac{\partial L_d}{\partial x_i}=x_i-m_{c_i}
\end{equation}

\textbf{Gradient with respect to $W_h$ ($\nabla W_h$):} For the parameter $W_h$, we give the chain rule as in equation \ref{equ:SRNETChain} considering the hallucination loss $L_h$ is added to the intermediate feature map $I_i^h$:
\begin{equation}
\label{equ:SRNETChain}
\frac{\partial L}{\partial W_h}=\Sigma_{i=1}^{n}\frac{\partial L}{\partial I_i^h}\frac{\partial I_i^h}{\partial W_h}
\end{equation} 

This shows we can run back propagation to get the gradient with respect to parameter set $W_h$ after we correct set the partial derivative of the loss function with respect to $I_i^h$. And by expanding $L$, we get:
\begin{equation}
\label{equ:SRNETChainExpand}
\frac{\partial L}{\partial I_i^h}=\alpha \frac{\partial L_h}{\partial I_i^h} +( \beta_1 \frac{\partial L_c}{\partial I_i^h}+\beta_2 \frac{\partial L_d}{\partial I_i^h})
\end{equation}

According to the definition, $\frac{\partial L_h}{\partial I_i^h}$ is quite simple as follows:
\begin{equation}
\frac{\partial L_h}{\partial I_i^h}=2(I_i^h-\tilde I_i^h)
\end{equation} 

And the remaining part, i.e.$( \beta_1 \frac{\partial L_c}{\partial I_i^h}+\beta_2 \frac{\partial L_d}{\partial I_i^h})$ can not be analytically expressed as $L_c$ and $L_d$ is not directly defined on $I_i^h$, however, it is just the result of back propagation of recognition layers.

\textbf{Gradient with respect to $W_s:$} This can be directly calculated according to the definition of softmax loss $L_c$.

\textbf{Center update $M$:} In deriving the gradient with respect to output feature $x_i$, we assume the center $m_j$ is fixed. However, during the training, $x_i$ will be inevitably changed, which requires to update $m_j$ accordingly. We strictly follow the mechanisms adopted in the literature \cite{centerloss} by updating the center $m_j$ with a learning rate $\gamma$ as it has been proven very effective:
\begin{equation}
m_j=m_j-\gamma \Delta m_j
\end{equation}
where $\Delta m_j$ is defined as:
\begin{equation}
\Delta m_j=\frac{\Sigma_{i=1}^{n}\delta(c_i=j) (m_j-x_i)}{1+\Sigma_{i=1}^{n}\delta(c_i=j)}
\end{equation}

With these gradient, we can easily run gradient descent algorithm iteratively to find the local minimum. We summarize the optimization algorithm in Algorithm \ref{alg:JointOptimization}:

\begin{small}
	\begin{algorithm}[htb]        
		\caption{Joint Optimization Algorithm}            
		\label{alg:JointOptimization}                
		\begin{algorithmic}[1]               
			\REQUIRE ~~\\                         
			Training samples $\mathcal{I}=\{<I_l^i,\tilde I_h^i,c_i>\}$;
			\ENSURE ~~\\                          
			Model parameter set $W=[W_h W_r]$
			\WHILE{not converged}
			\STATE t=t+1;
			\STATE calculate the partial derivative $\frac{\partial L}{\partial W_s}$;
			\STATE update the parameter set $W_s$ by $W_s^{t+1}=W_s^t-\theta \frac{\partial L}{\partial W_s}$;
			\STATE calculate the partial derivative $\frac{\partial L}{\partial x_i}$;
			\STATE execute back propagation from top layer to the bottom layer of FRNET to obtain $\frac{\partial L}{\partial W_r}$;
			\STATE calculate the partial derivative $\frac{\partial L_h}{\partial I_h^i}$;
			\STATE add the $\frac{\partial L_h}{\partial I_h^i}$ to the derivative $\frac{\partial L}{\partial I_h^i}$ obtained in step 6;
			\STATE execute back propagation from the top layer to the bottom layer of SRNET to obtain $\frac{\partial L}{\partial W_h}$;
			\STATE update the parameter W by $W^{t+1}=W^t-\theta \nabla$ W;
			\STATE calculate $\Delta m_j$;
			\STATE update the center $m_j$ by $m_j=m_j-\gamma \Delta m_j$;
			\ENDWHILE
		\end{algorithmic}
	\end{algorithm}
\end{small}

\section{Experiments}
In this section, we give the experimental results of our model. We first describe the experimental setting including the data preparation, network architecture and evaluation protocol. Then we give the performance of our models under different settings. Also, we compare performance of our SRNET with other state-of-art methods.

\begin{table*}[t]
	\centering
	\begin{tabular}[width=\textwidth]{|c|c|c|c|}
		\hline
		\thead{Layer type} & \thead{Kernel size} & \thead{Pad} & \thead{Stride} \\
		\hline
		Convolution & $9 \times 9, 64$ & $0$ & $1$ \\
		\hline
		Convolution & $1 \times 1, 32$ & $0$ & $1$ \\
		\hline
		Convolution & $1 \times 1, 3$ & $0$ & $1$ \\
		\hline
	\end{tabular}
	\caption{SRNET archicture details.}
	\label{tab:srnet}
\end{table*}

\begin{table*}[t]
	\centering
	\begin{tabular}[width=\textwidth]{|c|c|c|c|}
		\hline
		\thead{Layer type} & \thead{Kernel size} & \thead{Pad} & \thead{Stride} \\
		\hline
		Convolution & $3 \times 3, 64$ & $0$ & $1$ \\
		\hline
		Max Pooling & $2 \times 2$ & $0$ & $2$ \\
		
		\hline
		Residual & $\begin{bmatrix}
		3 \times 3, 64 \\
		3 \times 3, 64
		\end{bmatrix} \times 1 $ &
		$\begin{bmatrix}
		1 \\ 1
		\end{bmatrix} \times 1 $ &
		$\begin{bmatrix}
		0 \\ 0
		\end{bmatrix} \times 1 $ \\
		
		\hline
		Convolution & $3 \times 3, 128$ & $0$ & $1$ \\
		\hline
		Max Pooling & $2 \times 2$ & $0$ & $2$ \\
		
		\hline
		Residual & $\begin{bmatrix}
		3 \times 3, 128 \\
		3 \times 3, 128
		\end{bmatrix} \times 2 $ &
		$\begin{bmatrix}
		1 \\ 1
		\end{bmatrix} \times 2 $ &
		$\begin{bmatrix}
		0 \\ 0
		\end{bmatrix} \times 2 $ \\
		
		\hline
		Convolution & $3 \times 3, 256$ & $0$ & $1$ \\
		\hline
		Max Pooling & $2 \times 2$ & $0$ & $2$ \\
		
		\hline
		Residual & $\begin{bmatrix}
		3 \times 3, 256 \\
		3 \times 3, 256
		\end{bmatrix} \times 5 $ &
		$\begin{bmatrix}
		1 \\ 1
		\end{bmatrix} \times 5 $ &
		$\begin{bmatrix}
		0 \\ 0
		\end{bmatrix} \times 5 $ \\
		
		\hline
		Convolution & $3 \times 3, 512$ & $0$ & $1$ \\
		\hline
		Max Pooling & $2 \times 2$ & $0$ & $2$ \\
		
		\hline
		Residual & $\begin{bmatrix}
		3 \times 3, 512 \\
		3 \times 3, 512
		\end{bmatrix} \times 3 $ &
		$\begin{bmatrix}
		1 \\ 1
		\end{bmatrix} \times 3 $ &
		$\begin{bmatrix}
		0 \\ 0
		\end{bmatrix} \times 3 $ \\
		
		\hline
		Inner product & $512$ & - & - \\
		\hline
	\end{tabular}
	\caption{FRNET archicture details.}
	\label{tab:frnet}
\end{table*}

\begin{table*}[t]
	\centering
	\begin{tabular}[width=0.5\textwidth]{|c|c|c|c|c|}
		\hline
		\thead{Method} & \thead{Training images} & \thead{LFW Acc} & \thead{YTF Acc} \\
		\hline
		DeepFace \cite{DBLP:conf/cvpr/TaigmanYRW14} & 4M & $97.35\%$ & $91.40\%$ \\
		\hline
		DeepID-2+ \cite{DBLP:conf/cvpr/SunWT15} & - & $98.70\%$ & $-$ \\
		\hline
		FaceNet \cite{DBLP:conf/acl/FlekovaG16} & 200M & $99.63\%$ & $95.10\%$ \\
		\hline
		Center loss \cite{centerloss} & 0.7M & $99.28\%$ & $94.90\%$ \\
		\hline
		FRNET & 0.49M & $98.63\%$ & $91.30\%$ \\
		\hline
	\end{tabular}
	\caption{Verification performance of different methods on LFW and YTF datasets.}
	\label{tab:verif_performance}
\end{table*}

\begin{table*}[t]
	\centering
	\begin{tabular}{|c|c|c|c|c|c|c|}
		\hline
		\thead{Setting} & \thead{Training data} & \thead{Testing data} & \thead{LFW Acc} & \thead{LFW TP} & \thead{YTF Acc} & \thead{YTF TP} \\
		\hline
		1 & HR & HR & $98.63\%$ & $94.73\%$ & $91.30\%$ & $65.70\%$ \\
		\hline
		2 & HR & LR & $96.35\%$ & $74.66\%$ & $89.45\%$ & $46.10\%$ \\
		\hline
		3 & HR & Hallucinated & $96.30\%$ & $72.44\%$ & $89.36\%$ & $43.29\%$ \\
		\hline
		4 & LR & LR & $97.22\%$ & $82.40\%$ & $90.45\%$ & $61.20\%$ \\
		\hline
		5 & Hallucinated & Hallucinated & $97.61\%$ & $83.03\%$ & $88.20\%$ & $39.90\%$ \\
		\hline
		6 & LR & LR & $97.95\%$ & $88.73\%$ & $90.65\%$ & $58.50\%$ \\
		\hline
	\end{tabular}
	\caption{Accuracies and TPs in different settings.}
	\label{tab:settings}
\end{table*}

\begin{table}[t]
	\resizebox{\columnwidth}{!}{%
	\begin{tabular}{|c|c|c|}
		\hline
		\thead{Method} & \thead{Training data} & \thead{PSNR} \\
		\hline
		Bibubic & - & 30.08 \\
		\hline
		SRCNN & CASIA-WebFaces & 31.70 \\
		\hline
		Stand-alone SRNET & CASIA-WebFaces & 31.70 \\
		\hline
		Joint Model SRNET & CASIA-WebFaces & 31.71 \\
		\hline
	\end{tabular}
	}
	\caption{PSNR of different methods super-resolving LR-LFW}
	\label{tab:evalsr}
\end{table}

\subsection{Experimental Setting}
\textbf {Data Preparation}
We use 3 datasets in our experiments: CASIA-WebFace \cite{yi2014learning}, LFW \cite{LFWTech}, and YTF \cite{ytf}.
LR-CASIA, LR-LFW and LR-YTF are down-sampled versions of CASIA-WebFaces, LFW and YTF by a factor of $4$.
All the face images are aligned with 5 landmarks (two eyes, noise and mouth corners) detected with algorithm \cite{mtcnn} for similarity transformation. The faces are cropped to $124 \times 108$ RGB images.
Each pixel in RGB images is normalized by subtracting 127.5 then dividing by 128.
The only data augmentation we used is horizontal flipping.

\textbf{Network Architecture} This network consists of two parts: SRNET to hallucinate LR inputs and SRNET to extract deep discriminative features from input images. Details of SRNET and FRNET are given in Table \ref{tab:srnet} and Table \ref{tab:frnet}. The notation follows \cite{resnet}'s convention.

\textbf{Evaluation Protocol} We report our results on 3 metrics:
	1) Verification accuracy on LR-LFW and LR-YTF,
	2) True positive rate at low false positive rate $0.1\%$ (\textbf{TP} for short),
	and 3) Average PSNR gains on LR-LFW.

\textbf{Implemenation Details} We implement the SRNET and FRNET using the Caffe \cite{jia2014caffe} library with our modifications.
We extract the deep features by concatenating the output of the first fully-connected layer of the FRNET for each image and its horizontal flip. Verification task is done on the score computed by the cosine distance of two features after PCA.
For fair comparisons, we train the networks with batch size $128$. We choose a learning rate $0.00001$ for SRNET and a learning rate $0.1$ for FRNET, and divide the learning rates by $10$ after $16000$ and $24000$ iterations. The training procedure is finished after $28000$ epochs, in no more than $7$ hours on a single TITAN X GPU.

\subsection{Recognition Perfomance and Comparison}
One important goal of our model is to achieve better recognition performance for low resolution facial images. Thus we conduct the experiment using low resolution images for testing and compare with the methods that also use LR images as input.

\textbf{Setting 1: HR-training and HR-testing} In order to show the drop caused by low resolution images, we first give the recognition performance trained and tested by normal images, i.e. trained on CASIA-WebFaces and tested on LFW and YTF. For LFW testing set, the verification accuracy is $98.63\%$, and TP is $94.73\%$. For YTF testing set, the verification accuracy is $91.30\%$, and TP is $65.70\%$. We call the network trained by HR dataset as FRNET-HR. Also, we give a comparison of FRNET with other state-of-the-art models in Table \ref{tab:verif_performance}.

\textbf{Setting 2: HR-training and LR-testing} The simplest way to run face recognition for low resolution image is to directly feeding the up-scaled image into a network that is trained by the normal dataset (in our experiment, trained by HR-CASIA).  On LR-LFW testing set, we achieve accuracy $96.35\%$ and TP $74.66\%$. On LR-YTF, we achieve accuracy $89.45\%$ and TP $46.10\%$. We can see a large drop ($2.28\%$) on LFW compared with the number of using HR as inputs.

\textbf{Setting 3: HR-training and Hallucinated-testing} As SRNET produces hallucinated HR versions, we can also use the hallucinated images generated by the SRNET for testing. Thus we first train the SRNET using CASIA-Webfaces. By using the hallucinated versions of LR-LFW, we achieve verification accuracy $96.30\%$ on LFW, from which we can clearly see the hallucinated inputs even degrade the recognition performance compared with directly feeding the LR images to the network ($96.35\%$).

\textbf{Setting 4: LR-training and LR-testing} Another direct means to support LR testing is to train the network with LR-CASIA. We call this trained network FRNET-LR.
On LR-LFW testing set, we achieve accuracy $97.22\%$ and TP $82.40\%$. On LR-YTF, we achieve accuracy $90.45\%$ and TF $61.20\%$. FRNET-LR performs slightly better than FRNET-HR on LR versions of testing sets.

\textbf{Setting 5: Hallucinated-training and Hallucinated-testing} In order to directly benefit from the output of SRNET, we can train the FRNET by using the outputs of SRNET to improve the recognition performance. More precisely, we first train our SRNET and generate hallucinated version of LR-CASIA with SRNET, which are further used to train FRNET. In testing stage, we get the hallucinated versions of LR-LFW and LR-YTF and use the hallucinated versions for testing. Not 
surprisingly, we get accuracy $97.61\%$ on LR-LFW and $88.20\%$ on LR-YTF respectively. Surprising, It shows a improvement over previous settings on LR-LFW, and poses a negative impact to performance on LR-YTF.
We believe that the performance degradation on LR-LFW is caused by video compression artifacts which prevent the SRNET from working properly,
and more discriminative features can be learned from hallucinated face images to help recognition task.

\textbf{Setting 6: Joint End-to-end Training and Testing} In this setting, we give the recognition performance of our Joint Model. We train the network by taking LR-CASIA images as inputs and CASIA-WebFaces images as ground-truths. The weight of $\alpha$, $\beta_1$ and $\beta_2$ are set $0.01$, $1$ and $0.008$ respectively. We get accuracy $97.95\%$, TP $88.73\%$ on LR-LFW, and accuracy $90.95\%$, TP $59.40\%$ on LR-YTF, which shows a improvement over setting 5. Results of setting 5 and setting 6 support our hypothesis that not only FRNET can learn better from hallucinated images containing more discriminative features, but also SRNET can learn how to produce images more helpful to face recognition task.

We give accuracies and TPs under all 6 settings in Table \ref{tab:settings}.

\subsection{SR Performance and Comparison}

Our Joint Model serves not only for face recognition purpose, but also generates visually pleasing hallucinated images.
We trained a SRCNN from scratch as \cite{srcnn}, and compare it with our models.
Also, we find the Joint Model has slightly out-performanced stand-alone SRNET and SRCNN (trained on CASIA-WebFaces) by a $0.01$ dB.

\section{Conclusion}
In this paper, we have proposed a Joint Multi-tasking Model for LR face recognition and face SR.
By joining the SR network to our face recognition, the power of extracting deep feature from LR
is greatly enhanced. Experiments on several LR version of face benchmarks have convincingly demonstrated the
effectiveness of the proposed approach.

\bibliographystyle{plain}
\bibliography{egbib}

\end{document}